\documentclass[twoside,11pt]{article}

\usepackage{jmlr2e}

\ShortHeadings{Transforming the Output of GPT}{Ioste}
\firstpageno{1}

\begin{document}

\title{Transforming the Output of Generative Pre-trained Transformer: The Influence of the PGI Framework on Attention Dynamics}

\author{\name Aline Ioste \email ioste@ime.usp.br \\
       \addr Institute of Mathematics and Statistics\,
       University of São Paulo (IME-USP)\\
R. do Matão, 1010 - Butantã, São Paulo - SP, 05508-090 Brazil and \\
       \addr Department of Business Architecture, AI Specialist, 
       Itau Unibanco -Brazil}

\editor{}

\maketitle

\begin{abstract}
This paper presents a novel approach named Persona-Grouping-Intelligence (PGI), which has been crafted to tackle the challenges posed by GPT models when applied to real-world business issues. PGI leverages the inherent capabilities of the GPT model to comprehend intricate language structures and generate responses that are contextually relevant. The experiment occurred in a business scenario where human intelligence was being underutilized due to less optimized business processes. The primary objective of this approach is to leverage GPT models to reduce the workload on humans in tasks that are extensive, monotonous, and repetitive. Instead, the focus is redirected toward decision-making activities. Remarkably, the experiment yielded an accuracy rate of 93.81\% in validating 4,000 responses generated by the model, underscoring the effectiveness of the PGI strategies. Effectively addressing the issue of underutilized human intelligence, this paradigm shift aligns business environments with dynamic machine intelligence, enabling them to navigate the intricacies of real-world challenges. This approach facilitates the practical utilization of these models to tackle actual problems. The methodology offers an opportunity to reshape the fundamental structure of business processes by seamlessly integrating human decision-making with adaptable machine intelligence. Consequently, this optimization enhances operational efficiency and elevates strategic decision-making across diverse business contexts.
\end{abstract}


\section{Introduction}

The utilization of advanced language models, such as GPT (Generative Pre-trained Transformer) \citep{vaswani2017attention}, for automation within companies presents diverse challenges and necessitates a robust scientific foundation to mitigate risks and ensure the efficacy of these applications. While these models have demonstrated potential across various natural language processing tasks, they also bear notable limitations that could impact the reliability of decisions based on their outputs.

The emergence of LLM (Large Language Model) heralds a new era of automation and cognitive assistance, reshaping the interaction with information and decision-making processes \citep{stokel2023chatgpt}. The GPT model has garnered acclaim for its prowess in extracting information, showcasing a profound understanding of context, and delivering valuable insights across a multitude of subjects \citep{oh2023chatgpt}. This advancement marks a transformative shift in how automation and cognitive support are applied, offering the potential to redefine information engagement and decision-making through the capabilities of sophisticated language models.

Leveraging generalist models can yield significant advantages in automating specific company tasks; however, it concurrently introduces complexities and potential hazards that warrant a cautious and scientifically informed approach. The importance of a scientific foundation cannot be underestimated, as it plays a pivotal role in upholding the dependability, security, and effectiveness of such applications. This acknowledgment underscores the importance of meticulous consideration when applying these advanced language models in practical scenarios.

In most real business problems, companies cannot solely rely on the native knowledge of the GPT model but must augment it with external context to make informed and relevant decisions. This external context may include proprietary data, business-specific information, customer records, and other sensitive business-related data.

External context pertains to information or data that is provided to the language model alongside the prompt. In situations where information is absent from the training data of the model, the necessity of incorporating this new information necessitates the model to comprehend and adapt to unfamiliar contexts. This challenge can lead to potential inaccuracies or unintended outputs when the external context is not well-matched or appropriately handled.

To address these challenges, companies should implement robust data governance practices, invest in secure data management systems, and establish clear guidelines for using external context with the GPT model. Additionally, continuous monitoring and auditing of the performance of the model can help identify and mitigate potential biases or inaccuracies resulting from the use of external context. Making informed decisions about where to apply models as GPT requires careful consideration of data privacy, external context selection, business problem selection, and model calibration to ensure reliable and valuable results \citep{george2023review}.

The fundamental concept of prompts in natural language processing and artificial intelligence revolves around language models \citep{zhou2022large}. To effectively harness the power of language models like GPT in different scenarios, it is essential to gain a deeper understanding of the distinctive nature of prompts and the potential impact of external context when applied to real business problems.

To effectively utilize language models like GPT in different scenarios, developers and users must carefully consider the implications of external context and prompt nature, because, when considering the presence of an external context, the process becomes more complex compared to the context present within the training of the model corpus. 
This paper defines the nature of prompts as closed prompts and open prompts:

Closed prompts are meticulously designed and predefined within the codebase of the model for specific tasks. They offer a deterministic process, as the model directly processes the predefined input with minimal or no additional input required.

On the other hand, open prompts embrace a user-driven and interactive approach. Users have the freedom to craft prompts in their own words, presenting questions or context naturally. This fosters a dynamic and adaptable conversation with the model, enabling a wider range of queries and encouraging exploration. 

The choice between closed and open prompts depends on the nature of the task and the specific requirements of the business problem. Each approach has its advantages and limitations, and the decision should take into account the specific needs of each application scenario, as illustrated in Table \ref{tab:advantages_disadvantages}. Both approaches can be strategically employed to efficiently and accurately acquire the necessary information.

\begin{table}[htbp]
\centering
\caption{Advantages and Disadvantages of Closed and Open Prompts}
\label{tab:advantages_disadvantages}
\begin{tabular}{p{3cm}p{6cm}p{6cm}}
 & \textbf{Advantages} & \textbf{Disadvantages} \\
\textbf{Closed Prompts} & 
Precision and determinism: Closed prompts are predefined, leading to accurate and consistent responses in specific tasks. &
Limitation in flexibility: Closed prompts may be less adaptable to scenarios where more natural and open interactions are required. \\
& Control: Engineers have greater control over what the model should respond, ensuring more reliable results. & \\
\textbf{Open Prompts} & 
Flexibility: Open prompts allow for more natural and open interactions, enabling a wide variety of queries and creative responses. &
Less control: With open prompts, there may be less control over the responses of the model, resulting in more varied and, in some cases, less precise results. \\
& User engagement: Users have the freedom to express themselves in their own words, leading to a more engaging and interactive experience. & \\
& & Supervision complexity: As interactions are more open, more effort is required in supervising and correcting the responses of the model. \\
\end{tabular}
\end{table}

In the realm of business automation, where external context is incorporated due to data lying outside the GPT training of the model, it is critical to consider the risks associated with open prompts. These risks encompass ambiguous interpretations, unintended outcomes, information leaks, privacy concerns, and erroneous decision-making. Clarity and precision in instructions are crucial in business automation. Ambiguity or lack of detail in open prompts can lead to misunderstandings by the language model. Open prompts enable natural user interaction, yet this freedom comes with the risk of ambiguous or poorly structured queries, potentially resulting in incorrect or inadequate responses.

A recent study applied the GPT model to Stack Overflow (SO), a question-and-answer website for computer programming professionals. This research evaluated the responses generated by GPT for 517 questions. The study highlighted that GPT produced incorrect answers over half of the time, with 54\% of the errors attributed to the failure of the GPT in grasping question concepts \citep{kabir2023answers}.

Note that the questions in SO are authored by humans for humans, these questions frequently contain contextual gaps \citep{mondal2023automatic}. When introduced to models like GPT, the structure of such questions can result in a focus on irrelevant sections of the text. As a result, the generated solutions might be high-level but lacking a comprehensive grasp of the problem. It is important to note that machines lack a human-like understanding of context. A question that is clear to a human might not be as clear to a machine that relies on mathematical frameworks for interpreting context.

Considering our interaction with a non-conscious machine that relies on calculations for comprehension, the responsibility of directing the attention of the machine toward desired objectives falls on us, humans. Achieving this requires an in-depth understanding of effective attention-guiding techniques for such models. From the perspective of this paper, the SO study underscores the challenge of precisely directing machine attention more than it identifies a fundamental flaw in the linguistic model itself.

When dealing with a business problem, it is crucial that the outputs of the model align with the specific objectives and requirements of the company. In this scenario, open prompts heighten the risk of unexpected or unintended outcomes, which can have serious implications for business processes and results \citep{liu2023deid}. Due to the lack of strict constraints in open prompts, the model might generate contextually aligned but business-inappropriate responses, causing confusion or hindering the automation process \citep{rivas2023marketing}.

Furthermore, in business automation, privacy concerns are paramount as the system is likely to interact with sensitive data and confidential information. Open requests can lead to responses inadvertently exposing confidential information or raising security issues, potentially resulting in privacy breaches and reputational damage for the company \citep{demirci2022static}. Thoughtful management of interaction with the model is essential in business automation scenarios to prevent unintended data leaks and ensure data privacy \citep{ameri2021cybert}.

In business automation scenarios, the decisions made by the system can have a significant impact on business outcomes. If open prompts are not adequately controlled, the model may make incorrect decisions based on faulty assumptions or incorrect information, leading to unfavorable or harmful outcomes for the company \citep{he2023will}.

\section{Navigating External Context: Challenges in Expanding Attention Focus}

It is crucial to strike a balance between user flexibility and ensuring accurate and secure automated responses in the business automation context because both closed and open prompts in the context of business automation carry risks.  The risk of erroneous decision-making is even higher in open prompts, where users have more freedom to interact with the model \citep{deiana2023artificial}. The responses of the model may vary significantly based on how users phrase their queries, and without proper constraints, the model might generate responses that deviate from the desired outcomes.

To mitigate these risks, businesses must take them seriously and establish robust human curation and error-handling procedures. The outputs of the language model should be meticulously examined and validated before they are employed for decision-making purposes. By taking appropriate precautions and actively managing the risks, businesses can enhance the reliability and effectiveness of their automated systems and make more informed decisions based on the outputs of the model. By doing so, they can ensure the success of their business automation while minimizing the potential negative impacts of erroneous. 

Attention focus is a critical mechanism in language models, such as GPT, that allows them to weigh the importance of different words in the input text when generating responses. It enables the model to capture long-range dependencies and contextual information, making it capable of understanding and generating coherent and contextually relevant text \citep{ghojogh2020attention}.

Attention focus refers to the ability of the model to focus on specific parts of the input, influenced by the context or certain keywords in the prompt \citep{vig2019analyzing}.

In the case of closed prompts, where all the information is explicitly provided within the prompt itself, attention focus can be worked effectively. The model can concentrate on the tokens in the prompt, which are from its training corpus, and generate appropriate responses. However, the challenge arises when dealing with open prompts containing external context that is not part of the training data.

In open prompts with context not present in the training corpus, the model may face challenges in directing its attention to relevant information. This limitation of attention focus becomes particularly noticeable when the context involves specific or complex information.
Open Prompts with external context and attention-focus problems arise due to the limitations of the architecture of the transformer self-attention mechanism. While self-attention is effective in learning contextual representations within a text during pre-training, it faces challenges when confronted with unseen or external contexts not present in the training corpus \citep{wang2020linformer}.

The risky mix emerges as the attention focus mechanism faces challenges in effectively utilizing external context during inference. The mismatch between open prompts with external context and the attention focus in the transformer architecture can be exemplified.

The attention mechanism in language models is designed to focus on specific tokens or words in the input text to understand the context and generate relevant responses  \citep{wang2022shift}. However, if the external context contains data that is not part of the training corpus, it is considered "out-of-distribution" data. In such cases, the model may struggle to fully comprehend the meaning and relevance of the external context.

As a result, the attention of the model may be primarily directed toward the tokens in the prompt that it is familiar with, potentially neglecting or not fully considering the information from the external context. This can lead to suboptimal or erroneous decision-making, as the responses may not fully take into account the additional context provided in the external data.

To address this challenge, developers and users need to carefully design prompts and provide relevant and informative external context that aligns with the capabilities of the model.

\subsection{Enhancing Contextual Adaptation: Challenges and Strategies for Incorporating External Context}

Consider a situation where a language model is trained on a diverse corpus of general text data, and it is later employed to respond to queries in external contexts, such as a specific document. As these documents are not part of the training data, it faces an "out-of-distribution" scenario \citep{vaswani2017attention}.

To address this challenge and improve the performance of the model in incorporating external context, one approach is to train the language model on domain-specific data that includes content similar to the external context. By exposing the model to domain-specific information, it can adapt and become more effective in handling the new context. The training can be promising in various applications, enhancing the ability of the model to generate contextually relevant responses in domain-specific scenarios and offering more accurate and relevant responses in real-world applications \citep{alexandr2021fine}.

However, training a language model like GPT can be a resource-intensive task, both in terms of computational resources and human efforts. To perform training, a specific and relevant dataset for the target task is required. Obtaining, processing, and annotating this data can be a time-consuming and costly endeavor, particularly when dealing with substantial data volumes. Moreover, training demands significant processing time, and specialized resources like high-performance GPUs (Graphics Processing Units) or TPUs (Tensor Processing Units), which may not be readily available to everyone \citep{lee2020patent}.

For instance, just to understand the scale, a model with 355 million parameters enabled on a single GPU using 2.3 billion text tokens would take three months to train within 460,000 steps, setting a loss value of 2.6—this being a reasonable loss value as language models involve intricate grammatical constructs such as dependencies and structures that are not easy to capture \citep{khalil2022transforming}. It is estimated that GPT-4 has 170 trillion parameters \citep{openai2023gpt4}, which makes it significantly larger, more powerful, and more costly.


Without fine-tuning,  when utilizing the native GPT with external context incorporation, the prompt bears a significantly higher weight in terms of its necessity for effectively steering the attention of the model.

For example, consider a user who requests a native model like GPT with the query: \textit{ "Explain the potential legal responsibilities in the operational contract of a limited liability company"}. Without a clear and suitable context, as well as lacking precise guidance, the model might inadvertently provide coherent information about the subject matter, but not aligned with the expected user response due to the question allowing for various interpretations \citep{zhou2022large}.

The attention mechanism in transformers is key to understanding how prompts influence the model. When a prompt is given to the model, it is processed through the transformer layers, and during the self-attention step, the model calculates the relevance relationships between the words in the prompt and assigns weights to the most relevant information \citep{white2023prompt}.

Prompts direct the model, as the attention mechanism is trained to focus on the words or parts of the prompt that are most important for the given task. The quality and proper formulation of prompts are crucial to ensure that the model provides accurate and relevant responses. Designing effective prompts guides the model to focus on pertinent information, leading to improved results in natural language processing tasks \citep{zheng2023can}.

\section{ "Triple Bottom Line" in attentional guidance of models}

This paper defines the concept of "triple bottom line" in the attentional guidance of GPT models and refers to the approach of considering three interconnected aspects when customizing attention and generated responses by these models. These three aspects are analogous to the traditional three pillars of "triple bottom line" in sustainability, but adapted to the context of language models:

\begin{figure}[htb!]
\centering
\includegraphics[width=0.7\textwidth]{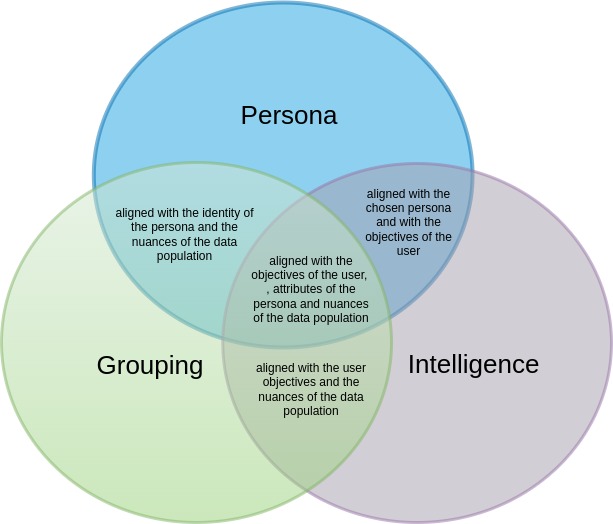}
\caption{PGI - The foundation of PGI rests upon three pivotal pillars, each meticulously designed to holistically guide the attention of the machine: Persona, Grouping, and Intelligence} \label{pci}
\end{figure}

\textbf{Persona:} The persona is the foundation of attentional guidance. It represents the identity, context, and characteristics that will steer the responses of the model. The persona serves as a lens through which the model interprets inputs and responds coherently in line with the defined personality and specifications.

\textbf{Grouping - Linguistic Context}: The linguistic context encompasses the environment in which the model operates. This involves understanding language nuances, contextual implications, and word relationships. Linguistic context is pivotal in generating responses that align semantically and grammatically with the presented input, which can be characterized by distinct groups of data samples. These groups can represent various contexts or scenarios within the input data.

\textbf{Intelligence - Human Intelligence Extraction:} User objectives represent the interests and needs of users interacting with the model. This includes grasping the purpose of the interaction, user expectations, and desired outcomes. Addressing user objectives is essential to delivering relevant and satisfactory responses.

Applying the "triple bottom line" concept to the attentional guidance of GPT models ensures that response personalization considers not only the persona but also the linguistic context and user objectives. This creates a balance among these three aspects, resulting in more meaningful, coherent, and user-centric interactions.

This novel approach named PGI (Persona-Grouping-Intelligence), figure \ref{pci}, has emerged as a means to mitigate the risks of attention of the model. The subsequent sections delve into the interplay of these three fundamental concepts.

\subsection{Persona-driven Attention: Enhancing Text Generation}

The concept of "Persona-driven Attention" involves attributing a specific role to the model, known as a persona. This persona encompasses characteristics, interests, and expertise relevant to a particular context. By incorporating this persona, the attention of the model is directed toward producing responses that align with the attributes of the persona. This results in a more coherent and relevant generated text.

The persona essentially acts as a filter through which the model processes input text. It helps the model understand the context and prompts, enabling it to generate responses that match the attributes of the persona. For example, if the persona is a legal expert, the model will focus on providing legal insights and information. This focused attention reduces the chances of generating unrelated or inaccurate outputs.

The persona serves as a contextual anchor, guiding the attention of the model and responses toward a specific domain. This enhances the ability to generate coherent, relevant, and accurate text tailored to a defined context or role.

The understanding of language model is facilitated by embedding dimensions that encode semantic and contextual information. These dimensions create a multi-dimensional space where each dimension corresponds to specific linguistic or contextual attributes. Assigning values to these dimensions allows the model to understand language and generate coherent text. 

Visualization techniques like Principal Component Analysis (PCA) can be used to simplify the understanding of the complex data relationships within the model. This is particularly useful due to the high-dimensional nature of the GPT operation. Visualizing data relationships in a lower-dimensional space aids in comprehending complex data structures.

This simplification allows the visualization of data relationships and patterns. This approach proves exceptionally beneficial for didactic purposes, as it facilitates comprehension of complex data structures. For example, in the text-embedding-ada-002, \href{https://platform.openai.com/docs/guides/embeddings/what-are-embeddings}{there exist 1,536 dimensions for embeddings}, making direct visualization challenging. (Information extracted as of the publication date of this article). 

\begin{figure}[htb!]
\centering
\includegraphics[width=\textwidth]{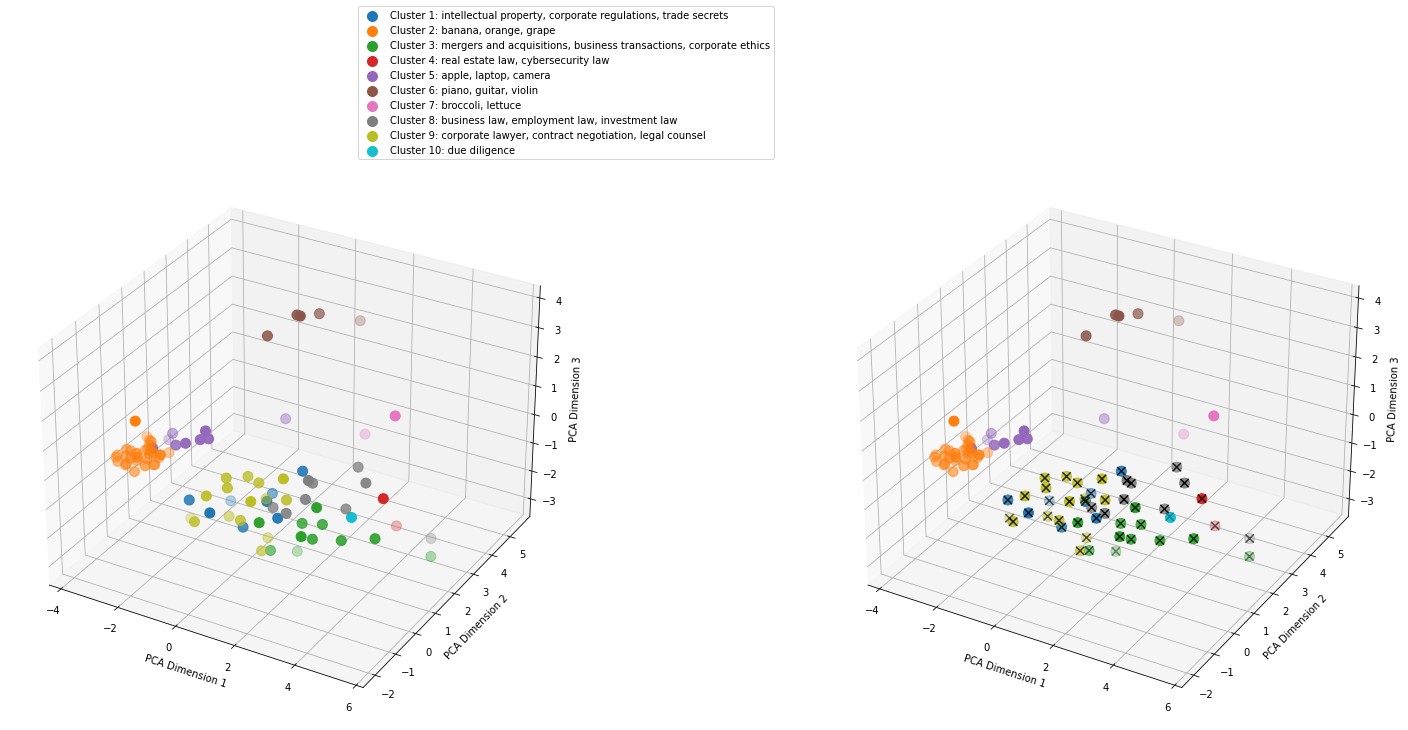}
\caption{Illustration of 100 term grouping using KMeans with 10 clusters and dimensionality reduction through PCA for visualization, distinguishing between persona-related and non-personal terms. The aim is to showcase semantic relationships and clusters among terms in a simplified 3D space using the BERT (Bidirectional Encoder Representations from Transformers) model.} \label{figembendding}
\end{figure}

In Figure \ref{figembendding}, the A graph displays terms grouped into clusters, each represented by distinct colors. Each cluster is labeled, showing three terms. The B graph presents terms marked with a black "X" denoting persona-related terms. This visualization provides insights into how embeddings semantically organize terms and their possible relationships to persona characteristics.

This term clustering analysis employs a three-dimensional space, where the objective is to visualize and better understand the grouping of terms related and unrelated to the persona.

This alignment aims to enhance the capacity of the model to draw informed conclusions. The GPT model has been trained across multiple languages. For precise interpretation of specific or technical documents, a strong grasp of the language in use is crucial for leveraging the inherent linguistic knowledge. Additionally, determining the nationality of the persona is a significant factor in accentuating unique aspects of the language the model is exposed to, given its training across multiple languages.

The model associates persona-specific information with corresponding linguistic structures, concepts, and relevant terminologies to become contextually relevant and tailored. This enables the model to better understand and interpret contexts, generating responses consistent with the expertise of the persona.

The persona can be utilized in both closed and open prompts, where its effectiveness is demonstrated in enhancing the text-generation process. In closed prompts, the persona functions as a filtering mechanism through which the model processes the input text. Similarly, in the case of open prompts, by assigning a persona to the model, users can guide the direction of the conversation and receive responses aligned with the expertise of the persona.

As the model tailors its responses to align with the attributes of the persona, it ensures that the generated text remains coherent and contextually relevant. In both closed and open prompt scenarios, the incorporation of persona-driven attention enhances the precision and adaptability of the text-generation process, leading to more valuable and relevant interactions.

The persona is a technical construct guiding the attention of the model toward a specific context. When a persona is employed, the model generates responses aligning with the role of the persona and knowledge. However, it is important to note that the model lacks consciousness and real experiences akin to human beings.

\subsection{Grouping - Linguistic Context}

In business scenarios, comprehending the attributes of the dataset is crucial for the exposure of the model. In the absence of this prior knowledge, data mining becomes essential to unveil patterns within this population, determining whether they are more concentrated or diverse. Grouping is employed here as a data extraction technique, grouping data points based on similarities. This aids in recognizing shared features, establishing a foundation to guide the focus of the model aligned with the attributes of the dataset population.

To enhance reliability and adaptability, a cluster-centered master prompt approach is employed. It categorizes groups based on shared attributes. This not only is used to improve precision but also supports controlled scalability, organizing the evaluation of the model to diverse types efficiently.

\begin{figure}[htb!]
\centering
\includegraphics[width=0.5\textwidth]{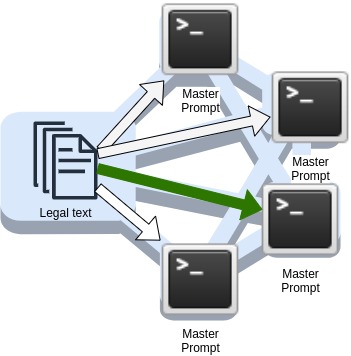}
\caption{Grouping of Prompt Sets by Contract Type} 
\label{figcluster}
\end{figure}

Each cluster leads to the creation of specialized master prompts, tailored to extract relevant information from the specific cluster. This strategy enhances accuracy and adaptability, enabling the model to effectively address diverse information requests. It also offers the potential for controlled expansion of the extraction of the model and comprehension capabilities, along with real-time monitoring possibilities. This aids in pinpointing clusters requiring increased human attention with greater accuracy.

This grouping approach is particularly valuable for categorizing the complexity of problems into parts. For instance, in the context of analyzing social contracts, even though categorized under the label "social contract," contracts can exhibit distinct complexities that require unique considerations at certain points. Some companies might produce documents that are simpler to analyze, whereas others present greater complexity due to specific regulations pertaining to delegation, representation, company management, and specific constraints regarding foreign partners.

Emphasizing the importance of noting that clustering should be conducted based on the type of data and further refined within each category according to its complexity. This methodology allows for a more detailed understanding of the intricacies presented by the problems the model will encounter.

Furthermore,  monitoring prompts inside each cluster through human curation is vital for maintaining model accuracy and effectiveness, especially in risk management. It provides previous insights acting as a risk mitigation layer. Reutilizing effective prompts and real-time monitoring identifies potential issues and optimizes pipelines.

This technique involves categorizing groups based on shared attributes, which is valuable for addressing the complexity of business problems.

In open prompt scenarios, clustering assists in refining questions and focusing on specific topics. By strategically categorizing clusters to match external contexts, businesses can tailor question scopes. For example, specific clusters tied to external contexts have context-adjusted master prompts with relevant questions. When users submit open prompts, clustering directs queries to relevant clusters, like channeling a lease-related question to the "Lease" cluster. Though not covering all context variations, clustering effectively shapes query scopes, enhancing responses.

In closed prompts, where all pertinent information is given, the cluster-centered master prompt approach significantly enhances reliability, adaptability, and response precision. Since closed prompts provide a clear context, clustering can guide the attention of the model and response generation. This ensures accurate and contextually relevant outputs, aligning with the specifics of the prompts. Thus, clustering optimizes response quality across both open and closed prompt contexts.

Clustering can find utility in scenarios involving both open and closed prompts. The strategic grouping of clusters based on various external contexts enhances the customization and quality enhancement of the responses generated by the model, irrespective of the type of prompt.

\subsection{Intelligence- Human Intelligence Extraction}

Human knowledge extraction, when integrated into the process, becomes a vital conduit to ensure the model generates coherent data aligned with user expectations. This highlights the importance of utilizing human expertise to shape the responses of the model, a core process within this strategy.

In the realm of model development, there is been a significant shift in focus. While in the past, developers primarily concerned themselves with the technical aspects such as the number of neurons in neural networks to ensure performance, the landscape has evolved. Now, attention is shifting towards a new paradigm that emphasizes the art of prompt engineering. It is no longer just about the technical architecture; it is about crafting prompts that effectively contextualize the understanding of the machine and guide its attention.

This transition signifies a fundamental change in approach. The success of a model is not solely determined by its architecture but by its ability to comprehend complex business contexts and generate coherent outputs aligned with specific goals. This calls for a holistic understanding of not just the technical intricacies, but also the intricacies of the business domain.

Enterprises need to realize that the integration of machine capabilities with the expertise of business specialists is paramount. The engineering of prompts has emerged as a crucial bridge between machine capabilities and business objectives. It is a collaborative process that involves a deep interaction with business experts. Extracting the nuances of their thought processes, understanding the intricacies of their decision-making, and grasping the context they operate within become pivotal steps in crafting prompts that drive coherent and aligned results.

This approach represents a fusion of technical prowess and domain knowledge. It is a dynamic interplay that is transforming the model development landscape. The focus has shifted from isolated technical decisions to a holistic strategy where machine learning experts work hand in hand with business specialists. This convergence ensures that models not only perform well technically but also generate outputs that are relevant, meaningful, and aligned with real-world business needs.

Human Intelligence Extraction is a vital factor in both open and closed prompt scenarios. For open prompts, where users formulate queries naturally, extracting intelligence relies on user input. Strategies include providing contextual clues, rephrasing queries, giving explicit instructions, engaging in iterative conversations, offering examples, and using keywords. These tactics guide the attention of the model and help users obtain relevant information from the model.

In closed prompt situations with explicit context, the collaboration between human expertise and machine capability is crucial. A language model specialist works to integrate business insights, extracting knowledge and structuring prompts to align responses with context. This process involves context, rephrasing, and sequencing instructions to maintain focus. Effective strategies ensure accurate responses, enhancing query relevance in the business domain.

\subsection{Interconnected Elements}

The PGI framework is built upon four interconnected elements, utilizing the interplay of "Persona", "Grouping", and " Intelligence". The elements can be implemented through four combinations: PG, GI, PI, and PGI as depicted in Figure \ref{pci}.

\textbf{PG:}This connection highlights the importance of aligning the linguistic context of the clustering with the  attributes of the persona. It enables the model to comprehend and respond appropriately within the context of the cluster while maintaining the characteristics of the persona.

\textbf{GI:}This connection focuses on harmonizing the cluster distribution with the objectives of the user. It ensures that the responses of the model not only align with the goals of the users but also reflect the cluster.

\textbf{PI:} This connection emphasizes the synergy between user objectives and the chosen persona. It ensures that the responses of the model are tailored to fulfill the intent of the user while embodying the traits of the persona, resulting in responses that are both purposeful and authentic.

\textbf{PGI:} This connection represents a harmonious collaboration between Persona, Grouping, and Intelligence, resulting in a comprehensive approach to maximizing the potential of GPT models. By seamlessly integrating user intent, persona attributes, and contextual insights, PGI redefines the landscape of human-AI interactions, offering purposeful, contextually accurate, and engaging responses that resonate with users and elevate the capabilities of AI in practical applications.

The four interconnections have specific roles in directing the attention of the model harmonizing the data population distribution, aligning with user objectives, and reflecting persona characteristics. Each of these interconnections addresses a different facet of the process and contributes to the overall quality of the outcome.
 
The PGI adjusts both the Attention Mechanism and Information Aggregation of the GPT model. By comprehending the advantages and challenges of these interconnected elements, a comprehensive framework is established. This framework empowers GPT models to generate responses that are not only contextually relevant but also align with the objectives. 

The practical application of these connections is explored in depth in the subsequent section, shedding light on the potential of the PGI framework in directing attention within GPT models.

\section{Experiment: Analyzing Legal Documents with GPT}

This experiment explored the capacity of generalist models like GPT to address a practical and real challenge within the business environment. Additionally, it validated the robustness of the framework PGI in guiding the attention of the model effectively, thereby mitigating the risk of responses that are misaligned with the objectives and expectations of the user.

\textbf{Exploring the Problem:} The business problem at hand revolves around the challenges faced by operations analysts who are responsible for navigating and understanding extensive legal documents, such as contracts. These documents often contain complex language, specific legal terms, and nuanced clauses that demand meticulous review. Due to the time-consuming nature of this task, analysts find themselves dedicating a significant portion of their time to reading and comprehending these documents. This, in turn, leaves them with limited capacity for higher-value activities, such as strategic decision-making.

The GPT model is employed to assist analysts in quickly comprehending the content of legal documents. By utilizing the ability of the model to process and interpret complex linguistic structures, the experiment seeks to provide a solution that can automate the extraction of key information, clauses, and insights from these documents. This automation can significantly reduce the time and effort required for manual document analysis.

The experiment aims to showcase the potential of the GPT model and the PGI (Persona-Grouping-Intelligence) methodology as a valuable tool for enhancing operational efficiency by automating the analysis of legal documents. By relieving analysts from the burden of extensive document review and providing them with targeted insights, they can allocate more time and attention to critical decision-making tasks, driving better business outcomes.

\textbf{The process:} The process initiates upon receiving contracts, followed by a data quality assessment of the document. Contracts meeting the minimum quality criteria undergo Optical Character Recognition (OCR) \citep{mori1999optical}, transforming scanned content into text. This textual data is then input into the GPT model for analysis. To ensure data confidentiality, the GPT model employed is an Itau private instance.

\begin{figure}[htb!]
    \centering
\includegraphics[width=\textwidth]{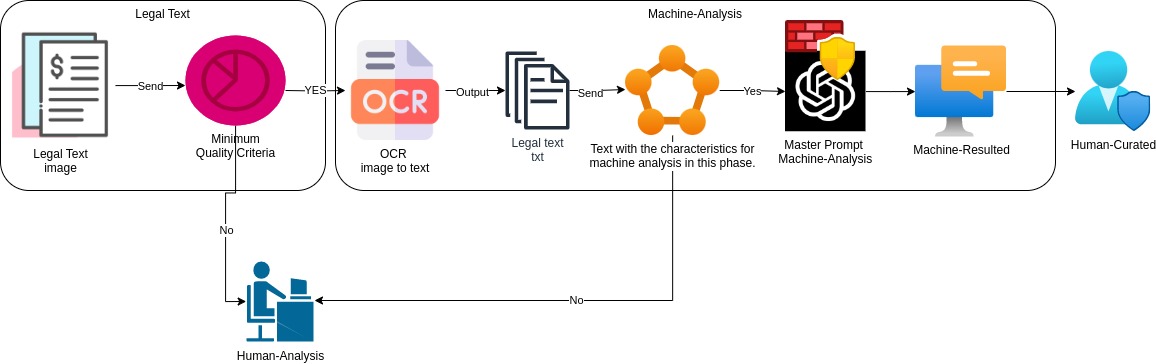}
\caption{Process analysis Legal Documents} \label{figapla}
\end{figure}

The model assesses 10 critical contract aspects essential for the operations decision-making of the team. The responses of the model regarding these points are relayed to the operations analyst for further evaluation, as depicted in Figure \ref{figapla}.

Recognizing the risks inherent in applying a generalist model in this context, the PGI methodology was rigorously implemented in this experiment. Specifically, all four connections of the "Triple Bottom Line" framework were utilized for attentional guidance of models. This comprehensive approach aims to bolster the attention of the GPT while reducing potential operational risks associated with process automation.

Each of the three elements was implemented in this experiment as follows:

\textbf{Persona:} A well-defined persona was created to embody a Brazilian lawyer specialized in business law. This persona was meticulously crafted to serve as a technical representation of the role of the expert, capable of deciphering the nuances of legal terms within social contracts, as well as understanding Brazilian legislation related to business law.

Incorporating Persona: \textit{"As a Brazilian lawyer specialized in business law, my task is to analyze the provided document and answer the following questions..."}

\textbf{Grouping:} A comprehensive analysis was conducted to categorize the diverse range of documents under review by the analysts. The findings revealed that 60\% of all analyzed documents were social contracts, with a monthly volume exceeding 10,000 contracts. With the aim of streamlining and enhancing the analysis process, each type of document examined by the operational team underwent a grouping procedure. This clustering process aimed to identify the pertinent information that needed to be extracted and examined from each distinct document category.

Given the substantial diversity within social contracts, a multi-level clustering approach was adopted. The primary clustering was further broken down into distinct subgroups. This stratification was crafted to offer more precise direction and guidance for different contract types, acknowledging the array of representative variations inherent within social contracts.

\begin{figure}[htb!]
\centering
\includegraphics[width=0.7\textwidth]{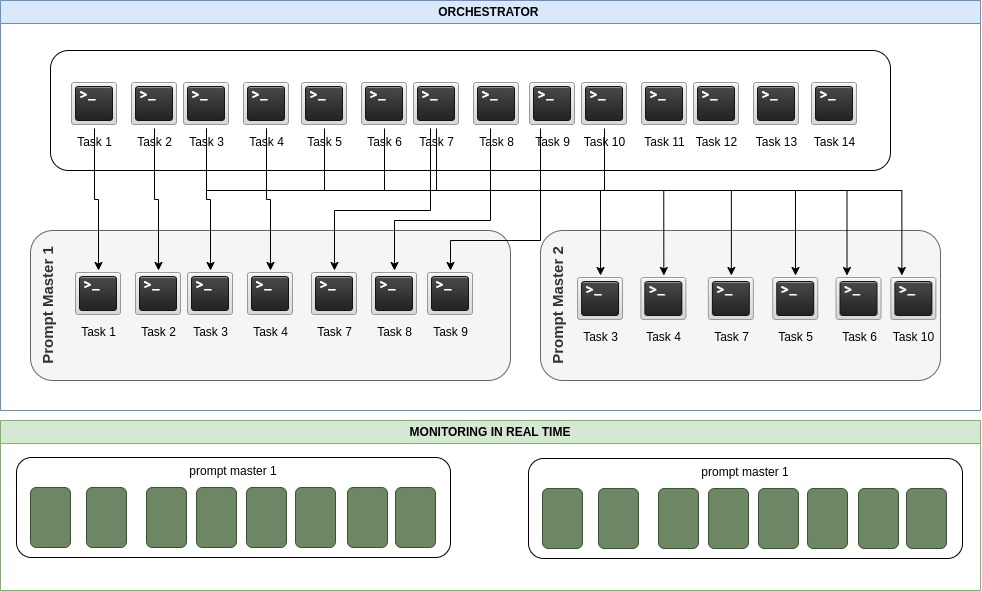}
\caption{Clustered Master Prompts and Real-time Monitoring: A visual representation showcasing the clustering of master prompts for distinct subgroups, each accompanied by corresponding gold-standard prompts. The real-time heatmap, fueled by the operational team, provides dynamic insights into model performance. It highlights successful model predictions and errors, enabling continual validation and enhancement} \label{monito}
\end{figure}

In response to the notable risks associated with decision-making and operational processes, a closed prompt strategy was implemented. A master prompt was created for each subcategory of the social contract. The formulation of these master prompts was informed by the analysts of the information necessary for decision-making within the contract. This encompassed both clause analysis and validation. The closed prompt approach was chosen to meet the rigorous demands of the operational context, effectively mitigating potential operational risks.

The grouping strategy adopted in this context was designed not only to manage the intricacies of various document types but also to facilitate the scalability of the evaluation of the capability machine across different document types, without affecting other groups. Additionally, real-time monitoring of machine-generated responses from the perspective of the operational analyst was introduced in the curation process. This approach aligns the outputs of the machine with the viewpoint of the operational analyst, ensuring responsive adjustments and continuous enhancement.

\textbf{Intelligence:} The master prompts are composed of a carefully crafted set of instructions, with each responsible for extracting specific information from the corresponding grouping they were designed for. During the prompt construction phase, there was close collaboration with domain experts who possessed specialized knowledge of the process and expertise aligned with the persona to which the model was personalized.

A close partnership was established with these experts, who held deep knowledge of the legal intricacies involved. Their detailed insights into legal nuances and complexities were extracted and refined to effectively contextualize the machine. These insights were transformed into prompts by IA experts to guide the model more precisely, ensuring that the machine could comprehend and respond adequately to the specific requirements of the legal domain.

The central objective of this approach was to empower the model to absorb the specialized knowledge of these professionals so that their guidance and instructions could be effectively conveyed. This not only enabled the machine to generate responses more aligned with human expectations but also ensured that the responses were legally accurate and contextually relevant. The direct collaboration with business and legal experts played a pivotal role in shaping the master prompts, making them effective tools for extracting significant information from the analyzed documents.

This deep interaction between technical and business experts enables AI to transcend the barrier of mere technical efficiency and become a true extension of business needs and objectives. After all, an AI that comprehends specific industry nuances and aligns with company values is an AI that provides tangible value. This symbiotic collaboration between technology and human expertise is the foundation upon which the next generation of AI is being constructed.

\subsection{Validation and Results}

To validate the effectiveness of this approach, an analysis of the distribution of the data population by contract type was conducted. A total of 400 social contracts belonging to a specific type were extracted. These contracts were then analyzed using the GPT model applied within the PGI strategy, consisting of four interconnected elements. This validation process involved generating a dataset of 4,000 model responses.

The model response validation team comprised five specialists with extensive expertise in contract analysis. Each of the 400 contracts was reviewed by these specialists, who evaluated the model-generated responses for accuracy and contextual relevance. Any inconsistency in the analysis was deemed an error, regardless of the correctness of the remaining response. Validation was only considered successful if the entirety of the response was accurate from the perspective of the evaluating specialist.

Furthermore, the validation by domain experts ensured that the results of the model adhered to the operational standards of the bank. Through a meticulous evaluation process, all 4,000 model-generated responses underwent thorough scrutiny, validating not only the GPT model itself but also the robustness of the PGI strategy in guiding the attention of the model.

The experimental endeavor, employing the developed model, yielded remarkable outcomes with an overall accuracy rate of 93.81\%. This achievement was attained through a comprehensive understanding of the functionality of generalist language models like GPT, enabling an influence on their Attention Mechanism and Information Aggregation via the PGI framework. The process of specialized and critical human validation played a pivotal role in ensuring result fidelity.

\begin{figure}[htb!]
\centering
\includegraphics[width=\textwidth]{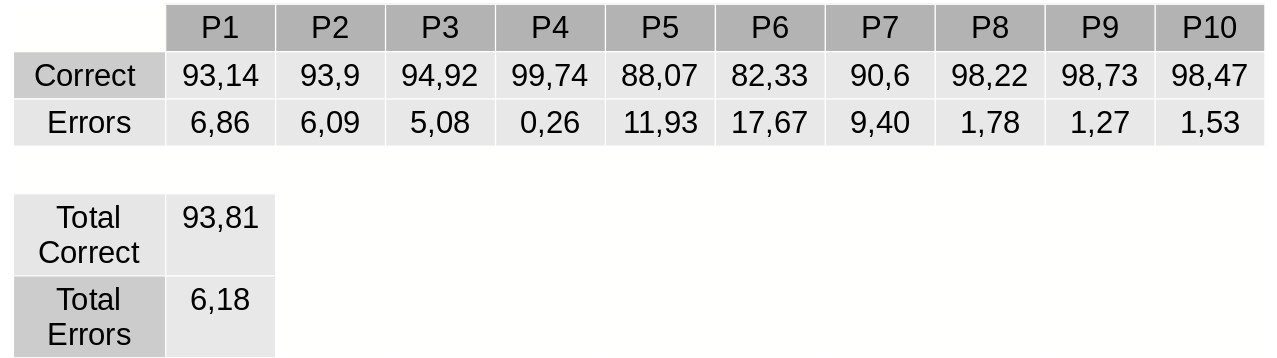}
\caption{Result of 4,000 GPT responses on ten tasks over 400 contracts} \label{result}
\end{figure}

The master prompts for cluster were formulated to analyze the social contracts from 10 perspectives within 400 contracts. For each of these 10 perspectives, the model demonstrated precision as detailed in Table \ref{result} in a total of 4.000 responses.

The model exhibited a performance that not only met but exceeded the expectations of experts in the domain to which GPT was applied, resulting in a mere 6.18\% error rate in the results. Upon thorough examination, the sources of the 6.18\% error instances were identified. 3.03\% of these errors were linked to inaccuracies in the OCR process, whereas only 3.15\% were connected to the performance of the GPT model.

This impressive level of precision serves as experimental evidence of the promising efficacy of the PGI concept in guiding model attention within the GPT framework. This strategic approach inherently recognizes the importance of contextual nuances and distinct persona attributes in situations where the model lacks specific domain knowledge, effectively enhancing the precision of the generalist model to address real-world business challenges.

\section{Conclusion}

In conclusion, this study introduces a paradigm shift in the domain of business process automation, showcasing the transformative capabilities of the PGI framework. By skillfully harnessing the capabilities of GPT within the PGI approach, the study illustrates the potential to redefine conventional business processes. This pioneering methodology capitalizes on the strengths inherent in human expertise and machine intelligence, fostering a harmonious collaboration that reshapes the business operations landscape.

The significance of the achieved results, along with meticulous validation, demonstrated an impressive accuracy rate of 93.81\%. The utilization of the PGI framework emerged as a pivotal strategy for deriving meaningful outcomes from Large Language Models (LLMs), which are susceptible to generating incoherent text when applied incorrectly. This experiment exemplified the robustness of the method in addressing the intricate challenge of legal text analysis, a task recognized for its complexity and time-intensive nature, within a real-world business context. This accomplishment lays the groundwork for organizations to embrace advanced AI technologies, reshaping their approach to intricate tasks, optimizing efficiency while upholding precision, and channeling human intelligence towards tasks of higher value.

The PGI technique emerges as a foundational driving force behind this transformative shift. Proficiently directing attention and establishing harmonious collaboration with these models signifies a pivotal stride toward guiding organizations toward data-informed decision-making. As we navigate a future where automation and AI assume increasingly pivotal roles, the capacity to harness the potential of language models like GPT and deploy them with strategic precision will assuredly distinguish companies within the competitive landscape. This transcends mere technological implementation; it is a strategic necessity that lays the groundwork for innovation, efficiency, and the astute allocation of human expertise to decision-making responsibilities. By adopting this approach, enterprises not only possess the opportunity to redefine their operational paradigms but also to unlock a realm of novel possibilities and untapped potential.


\acks{
Appreciation to Itaú Unibanco for its contribution to advancing generative AI research within the financial sector. Equally important is the commitment demonstrated by the organization in adhering to stringent data protection guidelines, thereby ensuring the integration of ethical considerations throughout the experimental process. The invaluable backing and resources provided by Itaú Unibanco throughout the course of the research significantly underscore the dedication exhibited by the institution. Furthermore, the provisioning of pertinent data and expert insights from the business domain has notably enriched the research findings, thereby amplifying their practical relevance within the landscape of real-world business complexities}

\newpage

\vskip 0.2in
\bibliography{sample}

\end{document}